\documentclass{article}
\pdfoutput=1
\usepackage{arxiv}
\setcitestyle{authoryear,open={(},close={)}}
\renewcommand{\cite}{\citep}

\usepackage[utf8]{inputenc} %
\usepackage[T1]{fontenc}    %
\usepackage{hyperref}       %
\usepackage{url}            %
\usepackage{booktabs}       %
\usepackage{amsfonts}       %
\usepackage{nicefrac}       %
\usepackage{microtype, color}      %

\usepackage{tabulary}
\usepackage{tabularx}
\usepackage{array}
\usepackage{algorithm}
\usepackage{algorithmic}
\usepackage{amsmath}
\usepackage{amssymb}
\usepackage{amsthm}
\usepackage{bbm}
\usepackage{graphicx}
\usepackage{indentfirst}

\usepackage{caption}
\usepackage{multirow}
\usepackage{makecell}
\usepackage{subfigure}

\usepackage{listings}

\newlength\savewidth

\makeatletter
\newtheorem*{rep@definition}{\rep@title}
\newcommand{\newrepdefinition}[2]{%
	\newenvironment{rep#1}[1]{%
		\def\rep@title{#2 \ref{##1}}%
		\begin{rep@definition}}%
		{\end{rep@definition}}}
\makeatother

\newrepdefinition{definition}{Definition}
\newrepdefinition{lemma}{Lemma}
\newrepdefinition{proposition}{Proposition}

\def\shownotes{1}  \ifnum\shownotes=1
\newcommand{\authnote}[2]{{[#1: #2]}}
\else
\newcommand{\authnote}[2]{}
\fi

\def\shownotes{0}  \ifnum\shownotes=1
\newcommand{\authornotenonurgent}[2]{{[#1: #2]}}
\else
\newcommand{\authornotenonurgent}[2]{}
\fi

\title{Unsupervised Few-shot Learning via Deep Laplacian Eigenmaps}

\author{%
	\large{Kuilin Chen} \\
	\large{University of Toronto}\\
	{\texttt{kuilin.chen@mail.utoronto.ca}} \\
	\And
	\large{Chi-Guhn Lee} \\
	\large{University of Toronto} \\
	{\texttt{cglee@mie.utoronto.ca}} \\
}
\ifdefined\usebigfont

\usepackage{times}
\usepackage[fontsize=13pt]{scrextend}
\AtBeginDocument{
	\newgeometry{left=1.56in,right=1.56in,top=1.71in,bottom=1.77in}
}
\else
\fi

\begin{document}
	
\maketitle
\begin{abstract}
	Learning a new task from a handful of examples remains an open challenge in machine learning. Despite the recent progress in few-shot learning,  most methods rely on supervised pretraining or meta-learning on labeled meta-training data and cannot be applied to the case where the pretraining data is unlabeled. In this study, we present an unsupervised few-shot learning method via deep Laplacian eigenmaps. Our method learns representation from unlabeled data by grouping similar samples together and can be intuitively interpreted by random walks on augmented training data. We analytically show how deep Laplacian eigenmaps avoid collapsed representation in unsupervised learning without explicit comparison between positive and negative samples. The proposed method significantly closes the performance gap between supervised and unsupervised few-shot learning. Our method also achieves comparable performance to current state-of-the-art self-supervised learning methods under linear evaluation protocol.
\end{abstract}

\section{Introduction}
Few-shot learning \citep{fei:etal:2006one} aims to learn a new classification or regression model on a novel task that is not seen during training, given only a few examples in the novel task. Existing few-shot learning methods either rely on episodic meta-learning \citep{finn:etal:2017,snell:etal:2017} or standard pretraining \citep{chen:etal:2019,tian:etal:2020rethinking} in a supervised manner to extract transferrable knowledge to a new few-shot task. Unfortunately, these methods require many labeled meta-training samples. Acquiring a lot of labeled data is costly or even impossible in practice. Recently, several unsupervised meta-learning approaches have attempted to address this problem by constructing synthetic tasks on unlabeled meta-training data \citep{hsu:etal:2019unsupervised,khodadadeh:etal:2019unsupervised,khodadadeh:etal:2021unsupervised} or meta-training on self-supervised pretrained features \cite{lee:etal:2021metagmvae}. However, the performance of unsupervised meta-learning approaches is still far from their supervised counterparts. Empirical studies in supervised pretraining show that representation learning via grouping similar samples together \citep{chen:etal:2019,tian:etal:2020rethinking,dhillon:etal:2020baseline,laenen:Bertinetto:2021} outperforms a wide range of episodic meta-learning methods, where the definition of similar samples is given by class labels. The motivation of this study is to develop an unsupervised representation learning method by grouping unlabeled meta-training data without episodic training and close the performance gap between supervised and unsupervised few-shot learning. 

\begin{figure}[ht]
    \vskip 0.2in
    \begin{center}
    \centerline{\includegraphics[width=\columnwidth]{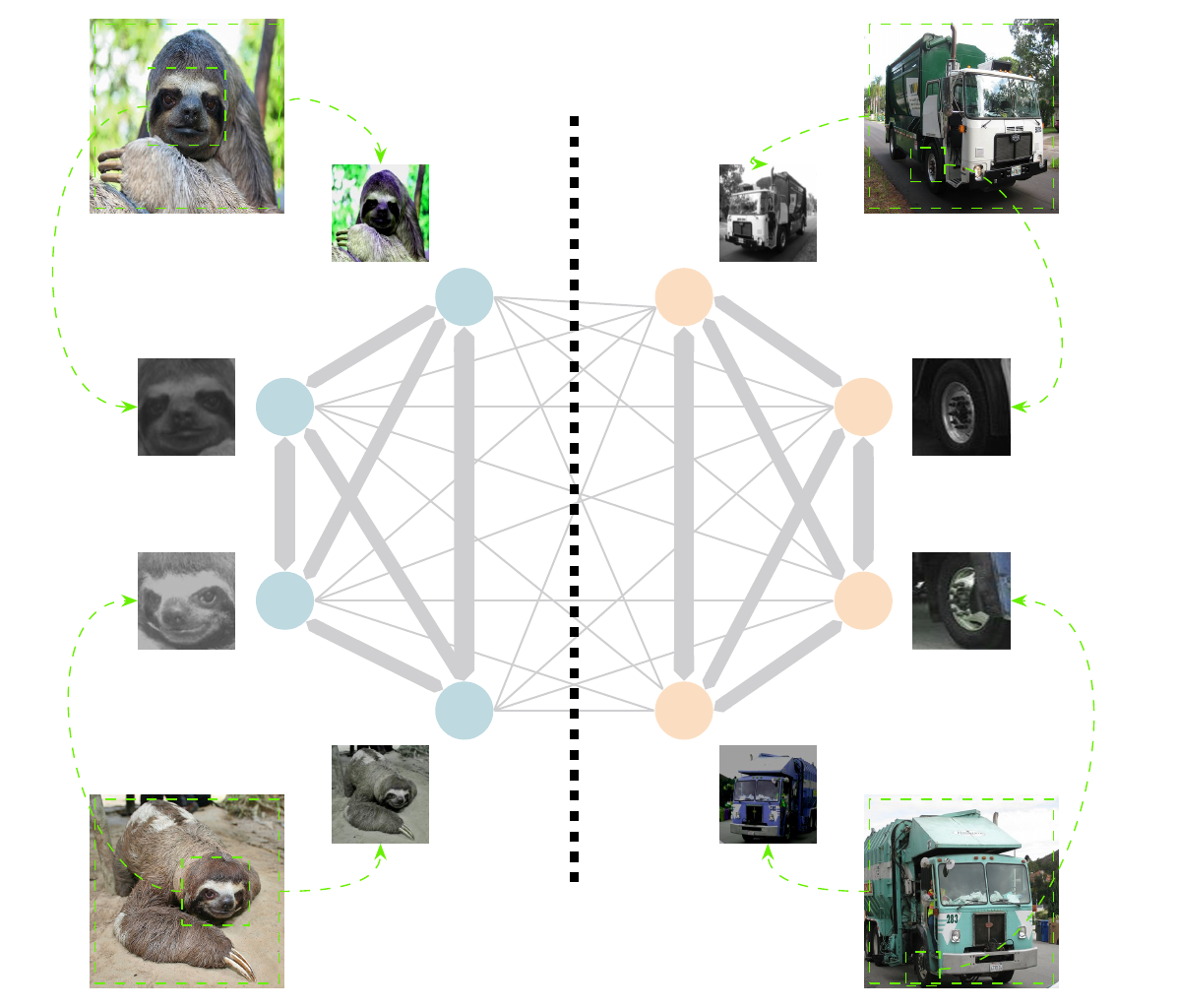}}
    \caption{A graph on augmented views of unlabeled training data. The thickness of the edge indicates the transition probability between vertices, which is proportional to their similarity. We group similar vertices together by minimizing the total transition probability between different groups.} \label{fig:graph}
    \end{center}
    \vskip -0.2in
\end{figure}

Contrastive self-supervised learning has shown remarkable success in learning representation from unlabeled data, which is competitive with supervised learning on multiple visual tasks \citep{tian:etal:2020contrastive,henaff:etal:2020cpcv2}. The common underlying theme behind contrastive learning is to pull together representation of augmented views of the same training sample (positive sample) and disperse   representation of augmented views from different training samples (negative sample) \citep{wu:etal:2018unsupervised,wang:isola2020}. Typically, contrastive learning methods require a large size of negative samples to learn high-quality representation from unlabeled data \citep{chen:etal:2020simclr,he:etal:2020moco}. This inevitably requires a large batch size of samples, demanding significant computing resources.  Non-contrastive methods try to overcome the issue by accomplishing self-supervised learning with only positive pairs. However, non-contrastive methods suffer from  trivial solutions where the model maps all inputs to the same constant vector, known as the collapsed representation. Various methods have been proposed to avoid collapsed representation on an ad hoc basis, such as asymmetric network architecture \citep{grill:etal:2020byol}, stop gradient \citep{chen:he2021simsiam}, and feature decorrelation \citep{ermolov:etal:2021whitening,Zbontar:etal:2021barlow,hua:etal:2021feature}. However, theoretical understanding about how non-contrastive methods avoid collapsed representation is limited, though some preliminary attempts are made to analyze the training dynamics of non-contrastive methods \citep{tian:etal:2021understanding}. Besides, most self-supervised learning methods focus on the linear evaluation task where the training and test data come from the same classes. They do not account for the domain gap between training and test classes, which is the case in few-shot learning and cross-domain few-shot learning.

We develop a novel unsupervised representation learning method in which a weighted graph is used to capture unlabeled samples as nodes and similarity among samples as the weights of edges. Two samples are deemed similar if they are augmentations of a single sample and clustering of samples is accomplished by partitioning the graph. We provide an intuitive understanding of the graph partition problem from the perspective of random walks on the graph, where the transition probability between two vertices is proportional to their similarity. The optimal partition can be found by minimizing the total transition probability between clusters. It is linked to the well-known Laplacian eigenmaps in spectral analysis \citep{shi:Malik:2000,meila:shi:2000learning,belkin:Niyogi:2003laplacian}. We replace the locality-preserving projection in Laplacian eigenmaps with deep neural networks for better scalability and flexibility in learning high-dimensional data such as images.

An additional technique is integrated into deep Laplacian eigenmaps to handle the domain gap between the meta-training and meta-testing sets. Previous studies on word embeddings  (e.g. king - man + woman $\approx$ queen) \citep{Mikolov:etal:2013} and disentangled generative models \citep{karras:etal:2019style} show that interpolation between latent embeddings may correspond to the representation of a realistic sample, which may not be seen in the training data. In contrast, interpolation in the input space does not result in realistic samples. In parallel, interpolation between the distributions of the nearest two meta-training classes in the embedding space can approximate the distribution of one meta-testing class after the feature extractor is trained \citep{yang:etal:2021free}. To enhance the performance on downstream few-shot learning tasks, we make interpolation of unlabeled meta-training samples on data manifold to mimic unseen meta-test samples and integrate them into unsupervised training of the feature extractor. 

Our contributions are summarized as follows:
\begin{itemize}
  \item A new unsupervised few-shot learning method is developed based on deep Laplacian eigenmaps with an intuitive explanation based on random walks.
  \item Our loss function is analyzed to show how collapsed representation is avoided without explicit comparison to negative samples, shedding light on existing feature decorrelation based self-supervised learning methods.
  \item The proposed method significantly closes the performance gap between unsupervised and supervised few-shot learning methods.
  \item Our method achieves comparable performance to current state-of-the-art (SOTA) self-supervised learning methods under the linear evaluation protocol.
\end{itemize}

\section{Methodology}

\subsection{Graph from augmented data}
First, we construct a graph using augmented views of unlabeled data. Let $\bar{\mathbf{x}} \in \mathbb{R}^d$ be a raw sample without augmentation. For image data, augmented views are created by the commonly used augmentations defined in SimCLR \citep{chen:etal:2020simclr}, including horizontal flip, Gaussian blur, color jittering, and random cropping. $\mathcal{X}$ denotes the set of all augmented data and $N=|\mathcal{X}|$. We represent the augmented data in the form of a weighted graph $\mathcal{G} = (\mathcal{X}, S)$, where each $\mathbf{x} \in \mathcal{X}$ is a vertex of the graph and $S$ denotes the edge weights. The edge between two vertices $\mathbf{x}_i$, $\mathbf{x}_j \in \mathcal{X}$ is weighted by the non-negative similarity $s_{ij}$ between them. For unrelated $\mathbf{x}_i$ and $\mathbf{x}_j$, the similarity $s_{ij}$ should be small. On the other hand, the similarity $s_{ij}$ should be large when $\mathbf{x}_i$ and $\mathbf{x}_j$ are augmentations from the same image or augmentations from two images within the same latent classes. An illustrative diagram is shown in Fig. \ref{fig:graph}.

Let $d_i = \sum_{\mathbf{x}_j \in \mathcal{X}} s_{ij}$ denote the total weights associated with $\mathbf{x}_i$, which is the degree of a vertex $\mathbf{x}_i$ in a weighted graph. The degree matrix $\mathbf{D}$ is defined as the diagonal matrix with the degrees $d_1, ..., d_N$ on the diagonal. The volume of $\mathcal{X}$ is $\mathrm{Vol}(\mathcal{X}) = \sum_{\mathbf{x}_i \in \mathcal{X}} d_i$. Similarly, the volume of a subset $C \subset \mathcal{X}$ is defined as $\mathrm{Vol}(C) = \sum_{\mathbf{x}_i \in C} d_i$.  Let $\mathbf{P}=\mathbf{D}^{-1} \mathbf{S}$ be the random walk Laplacian of $\mathcal{G}$, where $p_{ij} = (\mathbf{P})_{ij}$ represents the transition probability from $\mathbf{x}_i$ to $\mathbf{x}_j$, and each row of $\mathbf{P}$ sums to 1. $\mathbf{L = D -S} $ is the unnormalized Laplacian of $\mathcal{G}$.

\subsection{Random walks and graph partition}
$\mathcal{X}$ can be grouped into $K$ clusters, where similar vertices should be grouped into the same cluster and dissimilar vertices should be grouped into different clusters. It resembles the supervised pretraining by embedding samples from the same class together. We will show later that $K$ is also the dimension of the embedding. Since we do not know the number of classes in unlabeled training data, we set the embedding dimension $K=2048$, which works well on a wide range of datasets. Graph partition into clusters can be done by minimizing the total similarity $\sum_{\mathbf{x}_i \in C, \mathbf{x}_j \in C'} s_{i j}$ between two clusters $C, C' \in \mathcal{X}$, $C \cap C^{\prime}=\emptyset$. However, minimizing the total inter-cluster similarity is undesirable because it can simply separate one individual vertex from the rest of the graph. Instead, we run random walks on vertices $\mathcal{X}$. 

The random walk Laplacian $\mathbf{P}$ defines a Markov chain on the vertices $\mathcal{X}$. The stationary distribution $\pi$ of this chain has an explicit form $\pi_{i}=d_{i}/\operatorname{Vol}(\mathcal{X})$ for $\mathbf{x}_i \in \mathcal{X}$ \citep{meila:shi:2000learning}. The transition probability $P(C' \mid C) = P(X_1 \in C' \mid X_0 \in C)$ is given by 
\begin{equation}
  \begin{split}
  & P\left(X_{1} \in C' \mid X_{0} \in C \right) \\
  = & \left(\frac{1}{\operatorname{Vol}(\mathcal{X})} \sum_{\mathbf{x}_i \in C, \mathbf{x}_j \in C'} s_{i j}\right)\left(\frac{\operatorname{Vol}(C)}{\operatorname{Vol}(\mathcal{X})}\right)^{-1}\\
  = & \frac{\sum_{\mathbf{x}_i \in C, \mathbf{x}_j \in C'} s_{i j}}{\operatorname{Vol}(C)}
  \end{split}
\end{equation}
The inter-cluster transition probability is a proper criterion because it has a small value only if  $\sum_{\mathbf{x}_i \in C, \mathbf{x}_j \in C'} s_{i j}$ is small (low similarity for vertices in different clusters) and all clusters have sufficiently large volumes. As a result, minimization of the inter-cluster transition probability prevents trivial solutions that simply separate one individual vertex from the rest of the graph.

The inter-cluster transition probability minimizing problem is a constrained optimization problem. For the case of finding $K$ clusters, we define a matrix $\mathbf{Z} \in \mathbb{R}^{N\times K}$, where $z_{i k}$ represents the cluster assignment of the vertex $\mathbf{x}_i$. 
\begin{equation}
  z_{i k}=\left\{\begin{array}{ll}
  1 / \sqrt{\operatorname{vol}\left(C_{k}\right)} & \text { if } \mathbf{x}_{i} \in C_{k} \\
  0 & \text { otherwise }
  \end{array} \right.
\end{equation}
where $i=1, \ldots, N$ and  $k=1, \ldots, K$. Let $\mathbf{z}_{(k)}$ be the $k$-th column in the matrix $\mathbf{Z}$. The connection between unnormalized graph Laplacian and inter-cluster transition probability is given by
\begin{equation}
  \begin{aligned}
    & \mathbf{z}_{(k)}^{\top}\mathbf{Lz}_{(k)} =\mathbf{z}_{(k)}^{\top} (\mathbf{D} - \mathbf{S}) \mathbf{z}_{(k)} \\
  = & \sum_{i=1}^N d_i z_{ik}^2 - \sum_{i=1}^N \sum_{j=1}^N s_{ij} z_{ik}z_{jk} \\
  = & \frac{1}{2} \left( \sum_{i=1}^N d_i z_{ik}^2 - 2\sum_{i=1}^N \sum_{j=1}^N s_{ij} z_{ik}z_{jk} + \sum_{j=1}^N d_j z_{jk}^2\right) \\
  = & \frac{1}{2} \left( \sum_{i=1}^N \sum_{j=1}^N s_{ij} z_{ik}^2 - 2\sum_{i=1}^N \sum_{j=1}^N s_{ij} z_{ik}z_{jk} + \sum_{j=1}^N \sum_{i=1}^N s_{ji} z_{jk}^2\right) \\
  = & \frac{1}{2} \sum_{i, j=1}^{N} s_{i j}\left(z_{ik}-z_{jk}\right)^{2} \\
  = &\frac{1}{2} \sum_{i \in C_k, j \in \bar{C}_k} s_{i j}\left(\sqrt{\frac{1}{\operatorname{Vol}(C_k)}}\right)^{2}  = \frac{1}{2} P(\bar{C}_k \mid C_k)
  \end{aligned}
\end{equation}
It is easy to verify that $\mathbf{Z^{\top}DZ} = \mathbf{I}$, and $\mathbf{z}_{(k)}^{\top}\mathbf{Dz}_{(k)} = 1$. We can write the problem of minimizing inter-cluster transition as 
\begin{equation} \label{eq:original}
    \begin{aligned}
        \min _{C_{1}, \ldots, C_{K}} \quad & \operatorname{Tr}\left(\mathbf{Z^{\top} L Z}\right) \\
    \textrm{subject to} \quad & \mathbf{Z^{\top} D Z=I}\\
      &z_{i k} = 1/\sqrt{\operatorname{vol}\left(C_{k}\right)} \text { if } \mathbf{x}_{i} \in C_{k} \text { else } 0    \\
    \end{aligned}
    \end{equation}
This problem is NP-hard due to discreteness. Relaxing the discreteness condition, we obtain the relaxed problem
\begin{equation} \label{eq:relaxed}
  \min _{\mathbf{Z} \in \mathbb{R}^{N \times K}} \operatorname{Tr}\left(\mathbf{Z^{\top}} \mathbf{L } \mathbf{Z}\right) \text { subject to } \mathbf{Z^{\top} DZ=I}
\end{equation}
This is the standard trace minimization problem which is solved when the column space of $\mathbf{Z}$ is the subspace of the $K$ generalized eigenvectors of $\mathbf{Lz} = \lambda\mathbf{Dz}$. The relaxed problem in Eq. \eqref{eq:relaxed} leads to the lower bound on the optimal normalized cut of the graph \citep{chan:etal:1994spectral,zha:etal:2001spectral} and retains the interpretation of minimizing the transition probability between clusters.  In addition, such relaxation has asymptotic behavior when the number of data points tends to infinity \citep{luxburg:etal:2004limits}. As such, rounding $\mathbf{Z}$ leads to cluster indicator because the relaxed problem is a good proxy of the original problem in Eq. \eqref{eq:original} \citep{bach:Jordan:2006learning}.  We actually would not round the continuous $\mathbf{Z}$ as our goal is not clustering. We will learn a linear classifier on Z in the downstream tasks.

\subsection{Deep representation learning}
Let $\mathbf{z}_i \in \mathbb{R}^K$ be the $i$-th row of the matrix $\mathbf{Z}$. $\mathbf{z}_i$ can serve as desirable representation of $\mathbf{x}_i$ as it exhibits the clustering structure of the graph $\mathcal{G}$. However, it is not sensible to obtain $\mathbf{z}_i$ by generalized eigenvalue decomposition for two reasons. First, computation of eigenvectors may be prohibitively expensive due to the large size of augmented data $\mathcal{X}$. Second, it is non-trivial to compute the embeddings for unseen data points in meta-test classes because eigenvalue decomposition is non-parametric (one $K$-dimensional vector $\mathbf{z}$ is computed for each $\mathbf{x}$ in the training data). We assume that $\mathbf{z}$ is parametrized by $f_{\theta}(\mathbf{x})$, where $f_{\theta}$ can be deep neural networks with trainable parameters $\theta$. The relaxed problem in Eq. \eqref{eq:relaxed} is converted to 
\begin{equation} \label{eq:deep_eigenmaps}
  \min _{\theta} \operatorname{Tr}\left(\mathbf{Z^{\top}} \mathbf{L } \mathbf{Z}\right) \text { subject to } \mathbf{Z^{\top} DZ=I}
\end{equation}
where $\mathbf{Z} = f_{\theta}(\mathbf{X})$. We design a proper loss function to learn $\theta$ that solves the constrained optimization problem in Eq. \eqref{eq:deep_eigenmaps}.
The trace minimizing can be written as 
\begin{equation} \label{eq:weight_sum}
  \operatorname{Tr}\left(\mathbf{Z^{\top}} \mathbf{L } \mathbf{Z}\right) = \sum_{\mathbf{x}_i, \mathbf{x}_j \in \mathcal{X}} \left( s_{ij} \|f(\mathbf{x}_i) - f(\mathbf{x}_j)\|^2 \right)
\end{equation}
where the summation is taken w.r.t. pairs $(\mathbf{x}_i, \mathbf{x}_j)$ drawn i.i.d. from $\mathcal{X}$.  Note that the similarity $s_{ij}$ for an unrelated pair $(\mathbf{x}_i, \mathbf{x}_j)$ should be negligibly small, compared to the similarity of a related pair $(\mathbf{x}_i, \mathbf{x}_j)$. Therefore, the weighted summation in Eq. \eqref{eq:weight_sum} can be approximated by summation of the Euclidean distance between the representation of positive pairs within a mini-batch $\mathcal{V}_{\mathrm{batch}}$
\begin{equation} \label{eq:loss_sim}
  \mathcal{L}_{\mathrm{trace}} = \mathbb{E}_{\mathbf{z}, \mathbf{z}_+ \in \mathcal{V}_{\mathrm{batch}}}  \|\mathbf{z} - \mathbf{z}_+\|^2 
\end{equation} 
where $\mathbf{z}$ and $\mathbf{z}_+$ are representation of augmented views from the same image. 
The constraint $\mathbf{Z^{\top} DZ=I}$ requires the covariance matrix of the representation to be a diagonal matrix. It is equivalent to minimizing the mean squared error on off-diagonal entries of the covariance matrix
\begin{equation} \label{eq:loss_mse}
  \mathcal{L}_{\mathrm{const}}=\sum_{k=1}^K \sum_{l \neq k}^K (c_{kl})^{2}
\end{equation}
where $c_{kl} = \sum_{\mathbf{z} \in \mathcal{V}_{\mathrm{batch}}} (\mathbf{z})_k (\mathbf{z})_l/B$ is the covariance between the $k$-th and $l$-th dimensions of the feature, $B$ is the size of the mini-batch, and $(\cdot)_k$ denotes the $k$-th element of a vector.

The total loss is 
\begin{equation}
  \mathcal{L} = \mathcal{L}_{\mathrm{trace}} + \gamma \mathcal{L}_{\mathrm{const}}
\end{equation}
where $\mathcal{L}_{\mathrm{trace}}$ comes from trace minimization, $\mathcal{L}_{\mathrm{const}}$ is translated from the constraint on eigenvectors, and $\gamma$ can be treated as a Lagrange multiplier.

Our method does not require asymmetric twin neural networks, large batch size, large memory bank, or momentum update.  It naturally avoids the trivial solution via the orthogonality constraint, which is realized by decorrelating each dimension of the representation. In Section \ref{sec:decorrelation}, we will provide a more detailed analysis of how collapsed representation is avoided without explicit comparison between positive and negative samples in the loss function.

\subsection{Interpolation between unlabeled training samples}
\begin{figure}[h]
    \vskip 0.2in
    \begin{center}
    \centerline{\includegraphics[width=0.4\columnwidth]{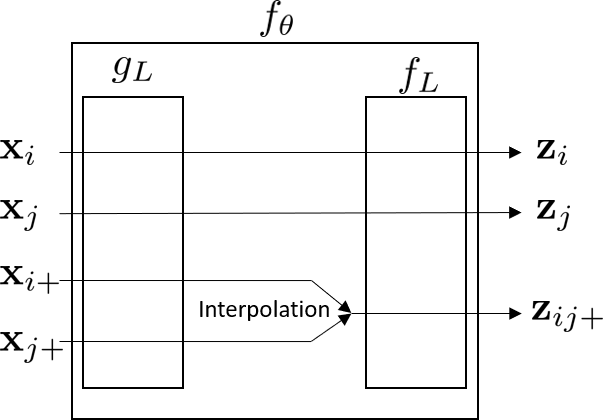}}
    \caption{Interpolation of unlabeled training data on data manifold.}
    \label{fig:mixup}
    \end{center}
    \vskip -0.2in
  \end{figure}

The representation of the augmented views can be encoded as $\mathbf{z} = f_{\theta}(\mathbf{x}) = f_L(g_L(\mathbf{x}))$, where $g_L$ is part of the feature extractor from the input layer to the hidden layer $L$ and $f_L$ is the remaining part of the feature extractor to the output layer.The hidden layer $L$ is randomly selected from a set of eligible layers in the neural network $f_{\theta}$ so that we can interpolate two samples on their intermediate representation. Let $\mathbf{x}_i$ and $\mathbf{x}_{i+}$ be a positive pair. We interpolate intermediate representation $g_L(\mathbf{x}_{i+})$ and $g_L(\mathbf{x}_{j+})$ to get the manifold mixup \citep{verma:etal:2019manifold} as follows
\begin{equation}
    \mathbf{z}_{ij+}=f_L(\lambda g_L(\mathbf{x}_{i+}) + (1-\lambda) g_L(\mathbf{x}_{j+})) 
\end{equation}
where $\lambda \in [0, 1]$ is a mixing coefficient drawn from a Beta distribution and $\mathbf{z}_{ij+}$ is the mixed representation of $\mathbf{x}_{i+}$ and $\mathbf{x}_{j+}$ after interpolation on the data manifold. $\mathbf{z}_{ij+}$ should match the interpolated representation of $\mathbf{z}_{i}$ and $\mathbf{z}_{j}$ in the embedding space with the same mixing coefficient. The trace minimization loss in Eq. \eqref{eq:loss_sim} can be replaced by 
\begin{equation} \label{eq:loss_sim_mix}
  \mathcal{L}_{\mathrm{trace}} = \mathbb{E}_{\mathbf{z}, \mathbf{z}_+ \in \mathcal{V}_{\mathrm{batch}}}  \|(\lambda\mathbf{z}_{i} + (1-\lambda)\mathbf{z}_{j}) - \mathbf{z}_{ij+}\|^2 
\end{equation} 
An illustrative diagram can be found in Fig. \ref{fig:mixup}. The pseudo-code is presented in Algorithm \ref{alg:our_algorithm}.

\begin{algorithm}[H]
    \caption{Pseudo-code of deep Laplacian eigenmaps in a PyTorch-like style.}
    \label{alg:our_algorithm}
    
     \definecolor{codeblue}{rgb}{0.25,0.5,0.5}
     \lstset{
       basicstyle=\fontsize{7.2pt}{7.2pt}\ttfamily\bfseries,
       commentstyle=\fontsize{7.2pt}{7.2pt}\color{codeblue},
       keywordstyle=\fontsize{7.2pt}{7.2pt},
     }
  \begin{lstlisting}[language=python]
    # B       : batch size 
    # K       : representation dim
    # d_x     : input dim
    # x       : Tensor, shape=[B, d_x]
    #           augmented views
    # x_p     : Tensor, shape=[B, d_x]
    #           positive pairs of x
    # z       : Tensor, shape=[B, K]
    #           representation of augmented views
    # z_p     : Tensor, shape=[B, K]
    #           positive pairs of z
    # gamma   : hyperparameter balancing the two losses
    # lambda  : interpolation coefficient
    # f_theta : feature extractor
    # L       : an randomly selected layer
    # g_L     : first L layers of f_theta
    # f_L     : remaining layers of f_theta
    # off_diag:  off-diagonal elements of a matrix
    
    def similarity(z, z_p):
        return (z - z_p).pow(2).mean()
    
    def decorrelation(z):
        c = z.T @ z / B
        decorr_loss = off_diag(c).pow(2).sum()
        
    z = f_theta(x)
    permuted_index = randperm(B)
    manifold_mix = lambda * g_L(x_p) 
          + (1 - lambda) * g_L(x_p[permuted_index])
    z_p_mix = f_L(manifold_mix)
    z_mix = lambda * z + (1 - lambda) * z[permuted_index]
    loss = similarity(z_mix, z_p_mix) + gamma * decorrelation(z)
  \end{lstlisting}
\end{algorithm}

\section{Unravel feature decorrelation} \label{sec:decorrelation}
We analyze the feature decorrelation loss in Eq. \eqref{eq:loss_mse} and show that feature decorrelation is indeed mathematically equivalent to contrasting between positive and negative samples. At first glance, the off-diagonal entries have nothing to do with the inner product between the representation of positive and negative samples. Previous works \citep{Zbontar:etal:2021barlow,hua:etal:2021feature} provide qualitative and empirical analysis to show that feature decorrelation avoids collapsed representation. To the best of our knowledge, we are the first to shed light on the equivalence and analyze the gradient of  the feature decorrelation loss.

When each dimension of $\mathbf{Z}$ is standardized to zero mean and unit variance (due to the final batch normalization layer), the diagonal entries of the covariance matrix $\mathbf{C}$ become constant $c_{ii} = 1$. Minimizing the square of off-diagonal entries in Eq. \eqref{eq:loss_mse} is equivalent to minimizing the Frobenius norm of the covariance matrix,
\begin{equation} \label{eq:trace}
  \begin{aligned}
    \|\mathbf{C}\|_F^2 & = \frac{1}{N^2}\|\mathbf{Z}^{\top}\mathbf{Z}\|_F^2 = \frac{1}{N^{2}} \operatorname{tr}\left(\mathbf{Z}^{\top} \mathbf{Z}\left(\mathbf{Z}^{\top} \mathbf{Z}\right)^{\top}\right)\\
  &=\frac{1}{N^{2}} \operatorname{tr}\left(\mathbf{Z} \mathbf{Z}^{\top}\mathbf{Z} \mathbf{Z}^{\top}\right) =\frac{1}{N^{2}} \operatorname{tr}\left(\left(\mathbf{Z} \mathbf{Z}^{\top}\right)^{\top} \mathbf{Z} \mathbf{Z}^{\top}\right)\\
  &=\frac{1}{N^{2}} \sum_{i=1}^N\sum_{j=1}^N\left(\left(\mathbf{Z} \mathbf{Z}^{\top}\right) \circ \left(\mathbf{Z} \mathbf{Z}^{\top}\right)\right)_{ij} \\
  &= \frac{1}{N^{2}} \sum_{i=1}^N\sum_{j=1}^N\left(f_{\theta}(\mathbf{x}_i)^{\top}f_{\theta}(\mathbf{x}_j) \right)^2
  \end{aligned}
\end{equation}
where $\circ$ denotes the Hadamard product. The second line of Eq. \eqref{eq:trace} is based on the cyclic property of the trace operation and the fact that $\mathbf{Z} \mathbf{Z}^{\top}$ is symmetric. After rewriting the trace operation via the Hadamard product, we have the final expression, showing that the squared Frobenius norm of the covariance matrix can be expressed as a summation over the squared inner product between pairs of representation. Note that $\mathbf{z}_i=f_{\theta}(\mathbf{x}_i)$ is usually projected to a sphere ball with a radius of 1 in self-supervised learning. The inner product between the representation of the same augmented view is constant $\mathbf{z}_i^{\top}\mathbf{z}_i=1$. Minimizing the square of off-diagonal entries in Eq. \eqref{eq:loss_mse} is equivalent to minimizing the total squared cosine similarity between random pairs.

After demystifying the feature decorrelation loss, we can analyze the gradient of our loss function to show how it pulls similar samples together and pushes dissimilar samples apart. Let $\mathbf{z}_i$ be the anchor sample, $\mathbf{z}_{i+}$ be a positive sample, and $\mathbf{z}_j$ be an unspecified sample in the current batch $\mathcal{V}_{\mathrm{batch}}$. The gradient with respect to the anchor sample is given by 
\begin{equation}
  \frac{\partial \mathcal{L}}{\partial \mathbf{z}_i} = \frac{2}{B}\left( \left( 1 - \gamma\right)\mathbf{z}_i - \mathbf{z}_{i+} + \gamma \sum_{\mathbf{z}_j \in \mathcal{V}_{\mathrm{batch}}} \frac{\mathbf{z}_i^{\top}\mathbf{z}_j}{B}\mathbf{z}_j  \right)
\end{equation}
where $B$ is the size of the mini-batch. Since $\gamma$ is a small positive number, the gradient can be further simplified as $\frac{\partial \mathcal{L}}{\partial \mathbf{z}_i} = \frac{2}{B}\left( (\mathbf{z}_i - \mathbf{z}_{i+}) + \gamma \sum_{\mathbf{z}_j \in \mathcal{V}_{\mathrm{batch}}} \omega_j\mathbf{z}_j  \right)$, where $\omega_j$ is a weighting factor for negative samples which is proportional to the similarity between the positive and negative samples. The first term of the gradient pulls positive pairs together while the second term disperses the negative samples.

\section{Related Work}
\textbf{Few-shot learning} is cast to optimization-based \citep{ravi:larochelle:2017,finn:etal:2017,nichol:etal:2018first,antoniou:Edwards:2018,lee:choi:2018gradient,park:Oliva:2019meta,flennerhag:etal:2020Warped,rusu:etal:2019meta,bertinetto:etal:2019meta} or metric-based \citep{koch:etal:2015siamese,vinyals:etal:2016,snell:etal:2017,qi:etal:2018,sung:etal:2018,oreshkin:etal:2018tadam,yoon:etal:2019tapnet,yoon:etal:2020xtarnet} meta-learning problems through supervised episodic training because it mimics the circumstances encountered in few-shot learning. The model is trained by a series of learning episodes, each of which consists of a limited number of support (training) samples and query (validation) samples. Nevertheless, simple baselines can outperform SOTA episodic meta-learning methods by using embeddings pre-trained with standard supervised learning \citep{chen:etal:2019,tian:etal:2020rethinking,dhillon:etal:2020baseline,mangla:etal:2020charting,laenen:Bertinetto:2021}. Although episodic meta-learning methods can be applied to unlabeled meta-training data by constructing synthetic tasks \citep{hsu:etal:2019unsupervised,khodadadeh:etal:2019unsupervised,khodadadeh:etal:2021unsupervised} or modeling the multi-modality within each randomly sampled episode \citep{lee:etal:2021metagmvae}, the performance is much worse than the supervised counterparts. Different from established few-shot learning methods, our method learns useful representation for downstream few-shot learning tasks using unlabeled meta-training data without episodic training. 

\textbf{Contrastive learning} with variants of InfoNCE loss \citep{gutmann:etal:2010nce,oord:etal:2018cpc} has been widely used in self-supervised/unsupervised representation learning \citep{wu:etal:2018unsupervised,henaff:etal:2020cpcv2,chen:etal:2020simclr,he:etal:2020moco}. It is derived from the maximization of the mutual information (MI) between related views \citep{poole:etal:2019variational}. However, this interpretation could be inconsistent with some empirical observations in self-supervised learning, such as tighter lower bounds of MI leading to worse performance \citep{mcallester:Stratos:2020,tschannen:etal:2020mutual,wang:isola2020}.  The InfoNCE loss can be expressed as $\mathcal{L}_{\mathrm{InfoNCE}} = -\sum_{i}\log\frac{\exp(\mathbf{z}_i^{\top}\mathbf{z}_{i+}/\tau)}{\sum_{\mathbf{z}_j\in\mathcal{V}} \exp(\mathbf{z}_i^{\top}\mathbf{z}_j/\tau)}$, where $\mathcal{V}$ can be a mini-batch or a memory bank. The core idea of contrastive learning is pulling positive pairs together while pushing negative samples apart \citep{wang:isola2020}. It can be easily verified from the gradient $\frac{\partial \mathcal{L}_{\mathrm{InfoNCE}}}{\partial \mathbf{z}_i} =  (-\mathbf{z}_{i+} + \sum_{\mathbf{z}_j\in\mathcal{V}} \omega_j \mathbf{z}_j)/\tau$, where $\omega_j = \frac{\exp(\mathbf{z}_i^{\top}\mathbf{z}_j/\tau)}{\sum_{\mathbf{z}_j\in\mathcal{V}} \exp(\mathbf{z}_i^{\top}\mathbf{z}_j/\tau)}$ is a weighting factor that is proportional to the similarity between the anchor sample $\mathbf{z}_i$ and the negative sample $\mathbf{z}_j$. Our method shares a similar form of the gradient with a different weighting scheme on negative samples, though our loss function is derived from a different perspective. Recent studies show that weighting schemes on negative samples affect the learning efficiency with respect to the negative sample size \citep{wang:liu:2021understanding,yeh:etal:2021decoupled}. Compared with contrastive learning with InfoNCE loss, our method does not require a large size of negative samples to work well.

\textbf{Clustering} methods have been employed in self-supervised learning \citep{caron:etal:2018deep,asano:etal:2019sela,caron:etal:2020swav} by simultaneously clustering the unlabeled data while enforcing consistent cluster assignments for different augmented views of the same image. These methods do not compare positive and negative samples directly as in contrastive learning, but careful implementation details and large batches are necessary because clustering methods are prone to collapse. The derivation of our method resembles spectral clustering but our method does not perform clustering. Unsupervised representation learning via deep Laplacian eigenmaps can be intuitively interpreted by random walks on augmented views of unlabeled data and use deep neural networks to handle high-dimensional image data. Random walks on image pixels have been developed to solve a supervised image segmentation problem by minimizing the Kullback-Leibler divergence between the learned transition probability and the target transition probability \citep{meila:shi:2000learning}. If $f_{\theta}$ is a linear function, the linear projection is locality preserving \citep{He:Niyogi:2004locality}.

\textbf{Feature decorrelation} methods avoid collapsed representation in self-supervised learning without using a large number of negative samples. Feature decorrelation can be achieved by differentiable Cholesky decomposition on each batch of embeddings \citep{ermolov:etal:2021whitening},  forcing the cross-correlation matrix of representations close to the identity matrix \citep{Zbontar:etal:2021barlow}, or utilization of decorrelated batch normalization with a shuffling operation \citep{hua:etal:2021feature}. Feature decorrelation methods show comparable performance to contrastive learning methods, but the fundamental principle behind it is unclear. Although Barlow Twins \citep{Zbontar:etal:2021barlow} are derived from the information bottleneck principle, it is only valid for Gaussian distributed data. The loss function in our method is similar to those in decorrelation methods. However, our method is derived from the well-known spectral analysis of the Laplacian matrix and requires minimal assumptions about the training data. Existing feature decorrelation methods in self-supervised learning can be unified in our framework - a trace minimization problem with orthogonality constraints, with minor differences in handling orthogonality constraints in practice. We also show the exact reason why feature decorrelation avoids collapsed representation.

\textbf{Mixup} \citep{zhang:etal:2018mixup} and its variants \citep{yun:etal:2019cutmix,verma:etal:2019manifold} provide effective data augmentation strategies to improve performance in supervised learning. Mixing up the image pixels has been explored in self-supervised learning \citep{shen:etal:2020unmix,lee:etal:2021imix}. MoChi \citep{Kalantidis:etal:2020} mixes the final representation of negative samples to create hard negative samples. Our method mixes up the intermediate representation to achieve better empirical performance. 

\section{Experiments}
We evaluate the performance of our model trained on unlabeled meta-training data on few-shot learning tasks, including in-domain and more challenging cross-domain settings. In addition, our method is also compared with SOTA self-supervised learning method under linear evaluation protocol to show that the proposed method can be applied to a wide range of downstream tasks beyond few-shot learning.

\subsection{Few-shot classification}
We conduct few-shot classification experiments on three widely used few-shot image recognition benchmarks.

\textbf{miniImageNet} is a 100-class subset of the original ImageNet dataset \citep{deng:etal:2009imagenet} for few-shot learning \citep{vinyals:etal:2016}. miniImageNet is split into 64 training classes, 16 validation classes, and 20 testing classes, following the widely used data splitting protocol \citep{ravi:larochelle:2017}.

\textbf{FC100} is a derivative of CIFAR-100 with minimized overlapped information between train classes and test classes by grouping the 100 classes into 20 superclasses \citep{oreshkin:etal:2018tadam}. They are further split into 60 training classes (12 superclasses), 20 validation classes (4 superclasses), and 20 test classes (4 superclasses).

\textbf{miniImageNet to CUB} is a cross-domain few-shot classification task, where the models are trained on miniImageNet and tested on CUB \citep{Welinder:etal:2010}. Cross-domain few-shot classification is more challenging due to the big domain gap between two datasets. We can better evaluate the generalization capability in different algorithms. We follow the experiment setup in \cite{yue:etal:2020}.

The feature extractor $f_{\theta}$ contains two components: a backbone network and a projection network. The backbone network can be a variant of ResNet architecture \citep{he:etal:2016}. The projection network is a 3-layer MLP with batch normalization and ReLU activation. The dimension of each layer in the projection MLP is 2048. We use the same augmentations defined in SimCLR \citep{chen:etal:2020simclr}, including horizontal flip, Gaussian blur, color jittering, and random cropping.

We use ResNet12 \citep{lee:etal:2019meta,ravichandran:etal:2019} and WRN-28-10 \citep{yue:etal:2020} as the backbone networks for few-shot learning and cross-domain few-shot learning, respectively. Those two backbone networks are selected because they are widely used in SOTA few-shot learning methods. The feature extractor is trained on unlabeled meta-training data by the SGD optimizer (momentum of 0.9 and weight decay of 5e-4) with a mini-batch size of 128. The learning rate starts at 0.05 and decreases to 0 with a cosine schedule. The projection network in the feature extractor is discarded after training on unlabeled data. 

During meta-testing, we train a regularized logistic regression model using 1 $\times$ 5 or 5 $\times$ 5 support samples on frozen representations after the global average pooling layer in the backbone network. Each few-shot task contains 5 classes and 75 query samples. The classification accuracy is evaluated on the query samples. 

\begin{table*}[!ht]
    \renewcommand\arraystretch{1.3}
    \centering
    \caption{Few-shot classification results on miniImageNet and FC100.}
    \scalebox{0.9}{
    \begin{tabular}{lcccccc}
    \hline
    \multirow{2}{*}{Method} & \multirow{2}{*}{Backbone}  & \multicolumn{2}{c}{miniImageNet 5-way}  & \multicolumn{2}{c}{FC100 5-way}\\
                                                      \cline{3-6} & & 1-shot & 5-shot & 1-shot & 5-shot  \\
    \hline
    Supervised                                 &           &                &  & & &\\ 
    Proto Net \citep{snell:etal:2017}          & ResNet-12 & $60.37 \pm 0.83$ & $78.02 \pm 0.57$ & $41.5 \pm 0.7$ & $57.0 \pm 0.7$ \\ 
    MAML \citep{finn:etal:2017}                & ResNet-12 & $56.58 \pm 1.84$ & $70.85 \pm 0.91$ & $36.9 \pm 0.6$ & $51.2 \pm 0.7$ \\ 
    TADAM \citep{oreshkin:etal:2018tadam}      & ResNet-12 & $58.50 \pm 0.30$ & $76.70 \pm 0.30$ & $40.1 \pm 0.4$ & $56.1 \pm 0.4$ \\
    Baseline++ \citep{chen:etal:2019}          & ResNet-12 & $60.83 \pm 0.81$ & $77.81 \pm 0.76$ & $41.3 \pm 0.7$ & $58.7 \pm 0.7$ \\
    MetaOptNet \citep{lee:etal:2019meta}       & ResNet-12 & $62.64 \pm 0.61$ & $78.63 \pm 0.46$ & $41.1 \pm 0.6$ & $55.5 \pm 0.6$ \\
    \hline
    Unsupervised                                 &           &                &  & & &\\ 
    CACTUs-MAML \citep{hsu:etal:2019unsupervised}   & ResNet-12 & $49.41 \pm 0.92$ & $63.72 \pm 0.83$ & $31.3\pm 0.8$ & $45.7 \pm 0.8$ \\
    UMTRA \citep{khodadadeh:etal:2019unsupervised} & ResNet-12 & $49.62 \pm 0.91$ & $62.43 \pm 0.84$ & $31.5 \pm 0.8$ & $45.3 \pm 0.8$ \\
    Meta-GMVAE \citep{lee:etal:2021metagmvae}  & ResNet-12 & $55.93 \pm 0.85$ & $74.28 \pm 0.72$ & $36.3 \pm 0.7$ & $49.7 \pm 0.7$ \\
    SimCLR \citep{chen:etal:2020simclr}        & ResNet-12 & $55.76 \pm 0.88$ & $75.59 \pm 0.69$ & $36.2 \pm 0.7$ & $49.9 \pm 0.7$ \\
    MoCo v2 \citep{he:etal:2020moco}           & ResNet-12 & $57.73 \pm 0.84$ & $77.51 \pm 0.63$ & $37.7 \pm 0.7$ & $53.2 \pm 0.7$ \\
    BYOL \citep{grill:etal:2020byol}           & ResNet-12 & $56.17 \pm 0.89$ & $76.17 \pm 0.66$ & $37.2 \pm 0.7$ & $52.8 \pm 0.6$ \\
    Barlow Twins \citep{Zbontar:etal:2021barlow} & ResNet-12 & $57.79 \pm 0.89$ & $77.42 \pm 0.66$ & $37.9 \pm 0.7$ & $54.1 \pm 0.6$ \\
    Ours                                       & ResNet-12 & $59.47 \pm 0.87$ & $78.79 \pm 0.58$ & $39.7 \pm 0.7$ & $57.9 \pm 0.7$ \\
    \hline
  \end{tabular}
  } 
  \label{tb:imagenet_resnet12}
\end{table*}

\begin{table}[ht]
    \renewcommand\arraystretch{1.3}
    \centering
    \caption{Cross-domain few-shot classification results on miniImageNet to CUB.}
    \scalebox{0.9}{
    \begin{tabular}{lcccc}
    \hline
    \multirow{2}{*}{Method} & \multirow{2}{*}{Backbone}  & \multicolumn{2}{c}{miniImageNet to CUB 5-way}  \\
                                                      \cline{3-4} & & 1-shot & 5-shot   \\
    \hline
    Supervised                                 &           &                  &  \\ 
    MAML \citep{finn:etal:2017}                & WRN-28-10 & $39.06 \pm 0.47$ & $55.04 \pm 0.42$  \\
    LEO \citep{rusu:etal:2019meta}             & WRN-28-10 & $41.45 \pm 0.54$ & $56.66 \pm 0.48$  \\ 
    MTL \citep{sun:etal:2019meta}              & WRN-28-10 & $43.15 \pm 0.44$ & $56.89 \pm 0.41$  \\
    Matching Net \citep{vinyals:etal:2016}     & WRN-28-10 & $42.04 \pm 0.57$ & $53.08 \pm 0.45$  \\
    SIB \citep{hu:etal:2020empirical}          & WRN-28-10 & $43.27 \pm 0.44$ & $59.94 \pm 0.42$  \\
    Baseline \citep{chen:etal:2019}            & WRN-28-10 & $42.89 \pm 0.41$ & $62.12 \pm 0.40$  \\
    \hline
    Unsupervised                                 &           &                  &  \\ 
    CACTUs-MAML \citep{hsu:etal:2019unsupervised} & WRN-28-10 & $33.48 \pm 0.49$ & $49.97 \pm 0.41$  \\
    UMTRA \citep{khodadadeh:etal:2019unsupervised} & WRN-28-10 & $33.59 \pm 0.48$ & $50.21 \pm 0.45$  \\
    Meta-GMVAE \citep{lee:etal:2021metagmvae}  & WRN-28-10 & $38.09 \pm 0.47$ & $55.65 \pm 0.42$  \\
    SimCLR \citep{chen:etal:2020simclr}        & WRN-28-10 & $38.25 \pm 0.49$ & $55.89 \pm 0.46$  \\
    MoCo v2 \citep{he:etal:2020moco}           & WRN-28-10 & $39.29 \pm 0.47$ & $56.49 \pm 0.44$  \\
    BYOL \citep{grill:etal:2020byol}           & WRN-28-10 & $40.63 \pm 0.46$ & $56.92 \pm 0.43$  \\
    Barlow Twins \citep{Zbontar:etal:2021barlow} & WRN-28-10 & $40.46 \pm 0.47$ & $57.16 \pm 0.42$  \\
    Ours                                       & WRN-28-10 & $41.08 \pm 0.48$ & $58.86 \pm 0.45$  \\
    \hline
  \end{tabular}
    }
  \label{tb:imagenet_cub}
\end{table}
  
The results of the proposed method and previous few-shot learning methods using similar backbones are reported in Table \ref{tb:imagenet_resnet12}. The proposed method outperforms existing unsupervised few-shot learning methods such as CACTUS \citep{hsu:etal:2019unsupervised} and UMTRA \citep{khodadadeh:etal:2019unsupervised} by a large margin. It demonstrates that high-quality representation for downstream few-shot tasks can be learned from unlabeled meta-training data without episodic training. Our method is in the category of self-supervised representation learning like SimCLR \citep{chen:etal:2020simclr}, MoCo v2 \citep{he:etal:2020moco}, BYOL \citep{grill:etal:2020byol}, and Barlow Twins \citep{Zbontar:etal:2021barlow} as it does not perform episodic learning. SimCLR achieves weaker performance than other self-supervised learning methods because it typically requires very large batch sizes to perform well. Although Meta-GMVAE \citep{lee:etal:2021metagmvae} performs unsupervised meta-learning on top of the pretrained features from SimCLR, the performance gain versus vanilla representation from SimCLR is at most marginal when deep backbones are used in our reproduction. This observation aligns with the empirical results in supervised few-shot learning where the advantage of episodic meta-learning diminishes as the backbone becomes deep \citep{chen:etal:2019,tian:etal:2020rethinking}. Our method is also compared with some strong baselines in supervised few-shot learning. The performance gap between supervised and unsupervised few-shot learning is significantly reduced by our method, compared with previous results in unsupervised few-shot learning \citep{hsu:etal:2019unsupervised,khodadadeh:etal:2019unsupervised,lee:etal:2021metagmvae}.
  
Our method is also applied to the cross-domain few-shot classification task as summarized in Table \ref{tb:imagenet_cub}. The proposed method outperforms other unsupervised methods in this challenging task, indicating that the learned representation has strong generalization capability. We use the same hyperparameters (training epochs, learning rate, etc.) from in-domain few-shot learning to train the model.  The strong results indicate that our method is robust to hyperparameter choice. Although meta-learning methods with adaptive embeddings are expected to perform better than a fixed embedding when the domain gap between base classes and novel classes is large, empirical results show that a fixed embedding from supervised or unsupervised pretraining achieves better performance in both cases. \cite{tian:etal:2020rethinking} also reports similar results that a fixed embedding from supervised pretraining shows superior performance on a large-scale cross-domain few-shot classification dataset. We still believe that adaptive embeddings should be helpful when the domain gap between base and novel classes is large. Nevertheless, how to properly train a model on unlabeled meta-training training to obtain useful adaptive embeddings in novel tasks is an open question.

\textbf{Ablation} studies are conducted to analyze how individual components affect the performance of few-shot learning. We study four variants of our methods: (a) the model is trained by only minimizing the similarity between positive pairs $\mathcal{L}_{\mathrm{trace}}$; (b) the projector network is a 2-layer MLP; (c) manifold mixup is not used in the model; (d) manifold mixup is replaced by input mixup. Table \ref{tb:cifar_alblation} shows the results of our ablation studies on FC100. When the model is trained without feature decorrelation, the accuracy on few-shot learning is close to random guess. It indicates that feature decorrelation is the key to avoiding trivial representation in learning from unlabeled data. After replacing the projector network with a 2-layer MLP, we can see obvious performance loss in the proposed method. Sufficient depth in the projector network is required to achieve optimal performance. The performance deteriorates without manifold mixup, indicating that manifold mixup helps the model to learn task-relevant information for downstream meta-test tasks. Compared with manifold mixup, input mixup is less effective in improving the few-shot learning performance.

\begin{table}[!ht]
    \renewcommand\arraystretch{1.3}
    \centering
    \caption{Ablation studies on FC100.}
    \scalebox{0.9}{
    \begin{tabular}{lcc}
    \hline
  \multirow{2}{*}{}    & \multicolumn{2}{c}{FC100 5-way}\\
                                                      \cline{2-3} & 1-shot & 5-shot  \\
    \hline
  Only $\mathcal{L}_{\mathrm{trace}}$               & Collapsed & Collapsed \\
  2-layer MLP                        & $36.1 \pm 0.7$ & $50.2 \pm 0.7$ \\
  Remove manifold mixup              & $38.2 \pm 0.7$ & $54.4 \pm 0.7$ \\
  Use input mixup                    & $38.7 \pm 0.7$ & $55.6 \pm 0.7$ \\
  \hline
  Ours                               & $39.7 \pm 0.7$ & $57.9 \pm 0.7$ \\
    \hline
    \end{tabular}
    } 
    \label{tb:cifar_alblation}
\end{table}

\subsection{Linear evaluation}
To examine the quality of the learned representation, we follow the linear evaluation protocol in self-supervised learning.  After the feature extractor is pretrained by unlabeled training data, a linear classifier is trained on top of the frozen backbone network using the labeled training data. The linear evaluation performance is widely used as the proxy for representation quality because it is highly correlated to the performance in downstream tasks, such as transfer learning, objection detection, and image segmentation \citep{he:etal:2020moco,chen:he2021simsiam}. Different from few-shot learning, unlabeled training data, labeled training data, and test data are from the same classes under the linear evaluation protocol.  We conduct experiments on CIFAR-10/100 and STL-10.

\textbf{CIFAR-10/100} are two datasets of tiny natural images with a size $32 \times 32$ \citep{krizhevsky:2009}. CIFAR-10 and CIFAR-100 have 10 and 100 classes, respectively. Both datasets contain 50,000 training images and 10,000 test images.

\textbf{STL-10} is a 10-class image recognition dataset for unsupervised learning \citep{coates:etal:2011stl10}. Each class contains 500 labeled training images and 800 test images. In addition, it also contains 100,000 unlabeled training images. Both labeled and unlabeled training images are used for feature extractor pretraining without using labels. The linear classifier is learned using the labeled training images.  

ResNet18 is adopted as the backbone network in the feature extractor. We train the feature extractor using SGD with momentum of 0.9 and weight decay of 5e-4. The learning rate starts at 0.05 and decreases to 0 with a cosine schedule. The feature extractor is trained for 800 epochs with a batch size of 256. 

After the feature extractor is pretrained by unlabeled data, a linear classifier is trained using SGD with a batch size of 256 and no weight decay for 100 epochs. The learning rate starts at 30.0 and is decayed by 0.1 at the 60th and 80th epochs. The test accuracy is reported in Table \ref{tb:linear_evaluation}. 

\begin{table}[!ht]
  \renewcommand\arraystretch{1.3}
  \centering
  \caption{Accuracy under linear evaluation protocol}
  \scalebox{0.9}{
  \begin{tabular}{lccc}
  \hline
          & CIFAR-10 & CIFAR-100 & STL-10  \\
  \hline
SimCLR          & 90.57  & 63.84 & 87.52  \\
MoCo v2         & 90.67  & 64.13 & 87.71  \\
BYOL            & 91.74  & 65.92 & 88.46 \\
Barlow Twins    & 91.58  & 65.83 & 88.65 \\
\hline
Ours            & 92.24  & 66.16 & 88.97 \\
  \hline
  \end{tabular}
  } 
  \label{tb:linear_evaluation}
\end{table}

Our approach achieves comparable performance to SOTA self-supervised learning methods in the linear evaluation under the same training recipe. Considering the good performance in linear evaluation, our method can be used in a wide range of downstream tasks beyond few-shot learning.

\section{Conclusions} \label{sec:conclusion}
In this article, we propose a new unsupervised few-shot learning method via deep Laplacian eigenmaps. Our method learns representation from unlabeled data by grouping similar samples together and can be intuitively interpreted by random walks on augmented training data. We provide a detailed analysis of our loss function derived from constrained trace minimization to show how it avoids collapsed representation analytically and the connection to existing self-supervised learning methods. The few-shot learning performance benefits from the interpolation of unlabeled training samples on the data manifold. Compared with existing unsupervised few-shot learning methods, the performance gap to supervised few-shot learning methods is significantly narrowed. Additional results on linear evaluation suggest that our method can be applied to a wide range of downstream tasks beyond few-shot classification.

\bibliographystyle{plainnat}
\bibliography{../bib/abbreviations,../bib/articles,../bib/proceedings,../bib/books}

\end{document}